\documentclass[11pt,a4paper]{article}
\usepackage[hyperref]{eacl2021}
\usepackage{times}
\usepackage{latexsym}

\usepackage{microtype}

\aclfinalcopy % Uncomment this line for the final submission
\usepackage{amsmath,amssymb}    

\newcommand{\personachat}{PersonaChat}
\DeclareMathOperator{\E}{\mathbb{E}}
\usepackage{mathtools}
\usepackage{booktabs}
\usepackage{tikz}
\usepackage{soul}
\usepackage{color}
\newcommand{\MM}[2][]{#2} % make it final draft

\newcommand{\MMF}[2][]{#2} % make it final draft

\title{Improving Factual Consistency Between a Response and Persona Facts}

\author{
\begin{tabular}{cccc}
Mohsen Mesgar & Edwin Simpson\thanks{* Now at the Intelligent Systems Lab, Dept. of Computer Science, University of Bristol.}
& Iryna Gurevych
\end{tabular}
\\
Ubiquitous Knowledge Processing Lab (UKP Lab) \\
Department of Computer Science, Technical University of Darmstadt \\
\url{https://www.ukp.tu-darmstadt.de}
}

\date{}

\begin{document}
\maketitle
\begin{abstract}
Neural models for response generation produce responses that are semantically plausible but not necessarily factually consistent with facts describing the speaker's persona. 
These models are trained with fully supervised learning where the objective function barely captures factual consistency. 
We propose to fine-tune these models by reinforcement learning and an efficient reward function that explicitly captures the consistency between a response and persona facts as well as semantic plausibility\footnote{\url{https://github.com/UKPLab/EACL21-personalized-conversational-system}}. Our automatic and human evaluations on the PersonaChat corpus confirm that our approach increases the rate of responses that are factually consistent with persona facts over its supervised counterpart while retaining the language quality of responses.  
\end{abstract}

\section{Introduction}
\emph{Response generation models} should ideally generate an appropriate \emph{response} to a given context consisting of utterances previously exchanged between dialogue partners and facts describing the speakers’ persona.  
These models have applications in developing dialogue systems as user interfaces for digital assistants~\cite{bobrow77} and also in asynchronous interactions in social media in which speakers define themselves by their profiles. 
\begin{table}[!t]
    \small
    \centering
    \begin{tabular}{@{}l@{}}
         \toprule
         \textbf{Persona}  \\
         \midrule 
         \begin{tabular}{l}
            \textit{fact 1}: i hate my job \\
            \textit{fact 2}: i ' m \underline{40 years old} \\
            \textit{fact 3}: i work as a car salesman \\ 
            \textit{fact 4}: my wife spends all my money  \\
            \textit{fact 5}: i am planning on getting a divorce 
        \end{tabular} 
        \\
        \midrule 
        \textbf{Dialogue History}  \\
        \begin{tabular}{p{7.5cm}}
            \textit{message}: hi , want to be friends?
        \end{tabular} 
        \\
        \midrule
        \textbf{Generated Responses}: \\
        \begin{tabular}{p{7.5cm}}
                \emph{inconsistent}: i ' d love to be friends . i ' m \underline{50 years old} \\
                \emph{consistent}: sure , \underline{i am 40} , i can tell you about myself
            \end{tabular}
        \\
        \bottomrule
    \end{tabular}
\caption{
A speaker's persona, dialogue history, one inconsistent, and one consistent possible response.  
}
\label{tab:sample_responses}
\end{table}

In this work, we focus on the aspects of persona that can be captured by a set of factual statements, a.k.a\MMF[]{.,} profiles. 
Table~\ref{tab:sample_responses}  illustrates the persona of the speaker who should respond to the given message. 
The first response is \emph{topically coherent with the message} and also \emph{linguistically fluent} (or in general, \emph{semantically plausible}) but \emph{factually inconsistent}, unlike the second response, with the second fact in the speaker's persona. 
We aim to improve the response quality in terms of its factual consistency with facts about the given speaker's persona while \MMF[to retain]{retaining} its semantic plausibility.  

Recent approaches to this problem~\cite{zhangsaizheng18,dinan19,wolf19} generate a response conditioned on persona facts and dialogue history and then use human-generated responses as demonstrations to train their models by fully supervised learning (SL).  
While this strategy has led to markedly improved performance, there is still a misalignment between this training objective -- maximizing the likelihood of human-written response\MMF[]{s} -- and what users care about -- generating semantically plausible and factually consistent outputs as determined by humans. 
This misalignment has several reasons: the maximum likelihood objective \MMF[has]{considers} no distinction between primary errors (e.g. inconsistent responses) and unimportant errors (e.g. selecting the precise word from a set of synonyms); models are incentivized to place probability mass on all human-generated responses, including those that are low-quality; and distributional shift during sampling can degrade performance. 
Optimizing for targeted quality factors is a principled approach to overcome these problems (e.g., \newcite{gao19} optimize text summarization systems for quality factors relevant to that task). 

Our goal is to advance methods for training response generation models on objectives that closely capture the behavior users care about. 
We first define a reward function to explicitly assesses the quality of a generated response according to \emph{factual consistency with persona facts}, \emph{topical coherence with dialogue history}, and \emph{language fluency}. 
We then train a policy via reinforcement learning (RL) to maximize the score given by our reward function; the policy generates a token of response at each ``time step'', and is updated using the \text{Actor-Critic} learning approach~\cite{mnih16} based on the ``reward'' our reward function gives to the entire generated response. 

We evaluate our approach on \personachat\ ~\cite{zhangsaizheng18}, a benchmark corpus of English dialogues designed to evaluate the factual consistency between a response and persona facts. 
We assess the language quality and the factual consistency of responses our RL-based model generates using automatic metrics and human evaluations
.  
Our core contributions are \MMF[two-fold]{twofold}:
\begin{itemize}
 \item We propose \MMF[]{to fine-tune a transformer-based response generation model by }an RL method including an efficient reward function that ensures factual consistency with persona facts as well as semantic plausibility of a response. 
    \item We use automatic and human evaluations to show that our RL-based method generates a response that is factually consistent with persona facts more frequently than its SL-based counterpart~\cite{wolf19}.     
\end{itemize}

The method we present in this paper is motivated in part by long-term concerns about the misalignment of NLP systems with what humans want them to do. 
When misaligned response generation models generate facts inconsistent with background knowledge like persona facts,  their mistakes are relatively low-risk and easy to catch. 
However, as these systems become more popular to solve essential tasks, their mistakes will likely become more subtle, making this an important area for further research. 

\section{Method}
\label{sec:method}
Let \mbox{$d=(u_1,...,u_{T-1})$} be the exchanged utterances between dialogue partners until turn $T-1$, and \mbox{$p= \lbrace f_1,..., f_{|p|}  \rbrace$} be a persona expressed by a set of facts (i.e. short sentences) about the speaker who should generate a response. Our goal is to generate a response \mbox{$r = (t_1,...,t_M)$} consisting of $M$ tokens so that $r$ is consistent with the facts in persona $p$, topically coherent with $u_{T-1}$, and linguistically fluent.

\subsection{TransferTransfo-SL}
\label{sec:deeprl}
We use the \emph{TransferTransfo} \cite{wolf19} dialogue model which is pre-trained and then \text{fine-tuned} with fully supervised learning (SL).  
\text{TransferTransfo} is a multi-layer transformer \cite{vaswani17} based on the Generative Pre-trained Transformer (GPT)~\cite{radford18}. Each transformer layer uses constrained self-attention where every token can only attend to its left context. 
Generation was performed using beam search with sampling, and an n-gram filtering is used to ensure the model does not directly copy from the persona facts nor former utterances. 
This model significantly improves over the traditional seq-to-seq, memory-based, and information-retrieval baselines in terms of (1)  topical coherence of the response, (2) consistency with a predefined persona, and (3) grammaticality and fluency as evaluated by the automatic metrics in the ConvAI2 competition~\cite{dinan19}. 
\MMF[]{Since this agent uses transformers, it copes with different lengths of dialogue history.}  

The transformer layers' parameters in this model are transferred from the pre-trained GPT and then are fine-tuned in a supervised scenario to optimize the losses for the response classification and response generation tasks. The former loss measures if the model distinguishes a correct response appended to the input sequence from a set of randomly sampled distractors, which are randomly selected. The latter one is the language modeling loss that measures how well the model can generate a response similar to the human-generated response. 
The generative loss is estimated as follows: the self-attention model's final hidden state is fed into an output softmax over the vocabulary to obtain the next response token probabilities. These probabilities are then scored using a negative log-likelihood loss, where the gold next tokens are taken as labels. 

\subsection{TransferTransfo-RL}
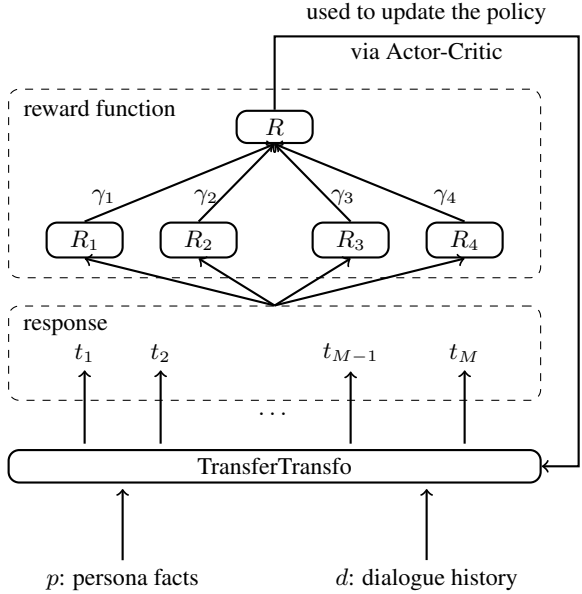
\begin{figure}
    \small
    \centering
    \begin{tikzpicture}

%\node[] (empty) at (0,0){};

\node[] (p) at (-2,1.5){$p$: persona facts};
\node[] (d) at (2,1.5){$d$: dialogue history};

%\node[draw,thick,rounded corners, minimum width=7cm] (gpt) at (0,1.5) {GPT};

\node[draw,thick,rounded corners, minimum width=7cm] (agent) at (0,3) {TransferTransfo};
%\node[draw,thick,rounded corners, minimum width=7cm] (agent) at (0,3) {Transformer-based Agent (Trans)};

\draw[->,thick] (p.north) -> (-2,2.7);
\draw[->,thick] (d.north) -> (2,2.7); 

%\draw[->,thick] (gpt) -> (agent);

\node[] (r1) at (-2.5,4.5){$t_1$};
\draw[->,thick] (-2.5,3.3) -> (r1);

\node[] (r2) at (-1.5,4.5){$t_2$};
\draw[->,thick] (-1.5,3.3) -> (r2);
%\draw[->,thick] (r1) |- (-2,4.5) -- (-2,3.7) ->(-1.5,3.7); 

\node[] (dots) at (0.0,3.7){$\dots$};

\node[] (rm-1) at (1.0,4.5){$t_{M-1}$};
\draw[->,thick] (1.0,3.3) -> (rm-1);
%\draw[->,thick] (0.5,3.7) ->(1.0,3.7); 

\node[] (rm) at (2.5,4.5){$t_{M}$};
\draw[->,thick] (2.5,3.3) -> (rm);
%\draw[->,thick] (rm-1) |- (2,4.5) -- (2,3.7) ->(2.5,3.7); 
\node[draw,dashed,rounded corners, minimum width=7cm,minimum height=1.25cm] (r) at (0,4.5) {};
\node[] (rlabel) at (-2.75,4.85) {response};

\node[draw,thick,rounded corners, minimum width=1cm] (R1) at (-2.5,6) {$R_1$};
\node[draw,thick,rounded corners, minimum width=1cm] (R2) at (-1,6) {$R_2$};
\node[draw,thick,rounded corners, minimum width=1cm] (R31) at (1.0,6) {$R_{3}$};
\node[draw,thick,rounded corners, minimum width=1cm] (R32) at (2.5,6) {$R_{4}$};
\draw[->,thick] (r.north) -> (R1.south);
\draw[->,thick] (r.north) -> (R2.south);
\draw[->,thick] (r.north) -> (R31.south);
\draw[->,thick] (r.north) -> (R32.south);

\node[draw,thick,rounded corners, minimum width=1cm] (R) at (0,7.5) {$R$};
\draw[->,thick] (R1.north) -> (R.south) node[pos=0.1, above] {$\gamma_1$};
\draw[->,thick] (R2.north) -> (R.south)node[pos=0.1, above] {$\gamma_2$};
\draw[->,thick] (R31.north) -> (R.south) node[pos=0.1, above] {$\gamma_{3}$};
\draw[->,thick] (R32.north) -> (R.south) node[pos=0.1, above] {$\gamma_{4}$};

\node[draw,dashed,rounded corners, minimum width=7cm,minimum height=2.5cm] (oracle) at (0,6.75) {};
\node[] (oraclelable) at (-2.30,7.75) {reward function};

\draw[->,thick] (R.north) |- (0,8.7) -- (4.0,8.7) -- (4.0,3) -> (agent.east); 
\node[] (ac) at (2,9) {used to update the policy};
\node[] (ac) at (2,8.5) {via Actor-Critic};
 
\end{tikzpicture}
    \caption{An abstract view of our RL approach.}
    \label{fig:method}
\end{figure}

Besides the remarkable improvement achieved by \text{TransferTransfo-SL}, its generated responses are not necessarily factually consistent with persona facts. For example, the inconsistent response in Table~\ref{tab:sample_responses} is generated by this system.     
We propose to fine-tune the parameters of this model using reinforcement learning (RL). 
The TransferTransfo model generates a response token-by-token for a given persona and dialogue history. 
After generating the last token, i.e.  `\texttt{<EOS>}',  or reaching the maximum length allowed for a response, a reward model assesses the quality of the response (Figure~\ref{fig:method}). 
The reward value is used to fine-tune the parameters of \text{TransferTranfo} towards the policy that generates a response that is factually consistent with persona facts and also semantically plausible. 

\paragraph{Action}
We consider generating each token of a response as an action performed by the \text{TransferTransfo} model:
\begin{equation}
    \mathcal{P}_\theta\left( r |  s \right) = \mathcal{P}_\theta(t_1|s) \prod_{k=2}^{M} \mathcal{P}_\theta(t_k|t_{1..k-1},s)\,\,\text{,}
    \label{eq:prob}
\end{equation}
where $t_k$ is the $k$th token in response $r$ and $t_{1..k-1}$ indicates the sequence of tokens generated prior to token $k$. 
For the sake of brevity, we use the notation $s$ to refer to $(p,d)$.  The function $\mathcal{P}_\theta(r|s)$ is the policy with the parameters $\theta$ of \text{TransferTransfo}.

\paragraph{Reward function}
A response generation system should ideally generate a response that is factually consistent with the persona facts, topically coherent with the former interactions, and linguistically fluent. 
Thus, we propose a compound reward consisting of four sub-rewards: $R_1$ ensures factual consistency with the persona facts.
$R_2$ accounts for topical coherence with the former utterance. $R_{3}$ and $R_{4}$ reinforce fluency. 
We use a weighted sum of these sub-rewards as the training signal: 
\begin{equation}
\label{eq:reward}
    R = \gamma_1 R_1 + \gamma_2 R_2 + \gamma_{3} R_{3} + \gamma_{4} R_{4}\,\,\text{,}
\end{equation}
where $\gamma_1 + \gamma_2 + \gamma_{3}  + \gamma_{4} =1$. 
These weights can be tuned as described below to prevent biasing the policy toward a particular \text{sub-reward}.  

\paragraph{Persona consistency sub-reward ($\mathbf{R_1}$)}
Recent studies \cite{welleck19,dziri19} show that consistency with factual information, such as persona facts, can be characterized as a natural language inference (NLI) problem, where  entailment labels can be taken as consistent labels and contradiction labels as inconsistent labels. 
Building on this, we use an NLI model to design this sub-reward.  
We define our NLI model using BERT as a bidirectional contextualized encoder: 
\begin{flalign}
   & h^{\texttt{[cls]}}, -  =  \texttt{BERT}(\texttt{[cls]} f_i \texttt{[SEP]} r)\,\,\text{,} & \nonumber \\
    & \left[ s_e, s_c, s_n \right]  = \texttt{MLP}(h^{\texttt{[cls]}})\,\,\text{,} & \\
    & \left[\mathcal{P}^{\text{NLI}}_e,\mathcal{P}^{\text{NLI}}_c,\mathcal{P}^{\text{NLI}}_n \right]= \texttt{Softmax}(\left[ s_e, s_c, s_n \right])\,\,\text{,} \nonumber &
\end{flalign}
where $f_i$ is a fact in the given persona, $r$ is the generated response, $\texttt{[SEP]}$ is the separator token, and $h^\texttt{[cls]}$ is  provided by BERT to classify semantic relationships between input  sentences~\cite{devlin2018bert}. $\texttt{MLP}$ is a linear layer 
that maps $h^\texttt{[cls]}$ to the scores $s_e$, $s_c$ and $s_n$, for the \emph{entailment}, \emph{contradiction}, and \emph{neutral} classes, respectively. $\mathcal{P}^{\text{NLI}}_e$, $\mathcal{P}^{\text{NLI}}_c$ and $\mathcal{P}^{\text{NLI}}_n$ denote the respective class probabilities.

We train our NLI model to predict the NLI classes of pairs of utterances and persona facts (\S\ref{sec:exp1}).  
We then use this trained model as $R_1$ to penalize the agent if its generated response contradicts one of the facts in the persona, and encourages the agent if its response entails a fact: 
\begin{equation}
    R_1 = \frac{1}{|p|}\sum_{f_i \in p}\mathcal{P}^{\text{NLI}}_e(f_i,r) - \frac{\beta}{|p|}\sum_{f_i \in p}\mathcal{P}^{\text{NLI}}_c(f_i,r)\,\,\text{,}
\end{equation}
where $\mathcal{P}^{\text{NLI}}_e$ and $\mathcal{P}^{\text{NLI}}_c$ are the entailment and contradiction probabilities of the relationship between $f_i$ and $r$. Scalar $\beta \geq 1$ is a marginal penalty for contradiction over entailment: responses that lack entailment may acceptably be neutral, while contradictory responses are a serious consistency error. 

The sub-reward for the factual consistency with persona facts is not sufficient to generate a semantically plausible response. The agent can maximize this sub-reward merely by repeating the persona's facts and ignoring topical coherence (for an example, see Appendix~\ref{sec:just-ps-subreward}). To prevent such behavior, we assess the topical coherence and grammatical fluency of a response by the following sub-rewards.  

\paragraph{Topical coherence sub-reward ($\mathbf{R_2}$)}
Topical coherence is a crucial property of high-quality dialogues \cite{see19,mesgar-etal-2020-dialogue}. 
We capture the topical coherence of response $r$ to the last utterance $u_{T-1}$ in dialogue history by representing them using an average pooling layer over their token representations obtained by BERT. 
Inspired by~\newcite{baheti18} and~\newcite{see19}, we use \emph{cosine similarity} between $\vec{r}$ and $\vec{u}_{T-1}$ as a proxy for topical coherence: 
\begin{equation}
    R_2= \texttt{cos}(\vec{r},\vec{u}_{T-1})\,\,\text{.}
\end{equation}

\paragraph{Fluency sub-rewards ($\mathbf{R_{3}}$ and $\mathbf{R_{4}}$)}
The above sub-rewards do not assess if the response content expressed is linguistically fluent. 
As also suggested in prior work \cite{yarats18,zhaotiancheng19,baosiqi19}, applying RL for specific metrics  might bring in adverse impacts on linguistic quality. 
As such, we add sub-rewards $R_3$ and $R_4$ to promote linguistic quality.   
$R_3$ employs a language model (LM) fine-tuned on a set of utterances (\S\ref{sec:exp2}) to evaluate the language quality of response. 
To do so, we use the Negative Log-Likelihood (NLL) loss obtained by this LM: 
\begin{equation}
 R_{3} =  \frac{\alpha - \texttt{NLL}(r)}{\alpha},       
\end{equation}
where parameter $\alpha$ is used to map any value of NLL that is greater than $\alpha$ to $\alpha$ so that the output of $R_3$ will be between $0$ and $1$. 
To retain the language quality of responses similar to those of TransferTransfo, we set  $\alpha$ to the maximum NLL value that this LM returns for responses generated by the TransferTransfo model on a development set. 
$R_3$ is not biased to the length of a response as NLL is already normalized by response length. 

Repeated tokens in a response \emph{significantly and negatively} influence the quality of the response~\cite{see19}. $R_4$ specifically discourages the generation of 1-gram tokens that appear in a response more than one time in a row: 
\begin{equation}
    R_{4} = 1- \frac{\# \text{repeated-tokens-in-response}}{\# \text{tokens-in-response}}.
    \label{eq:wr}
\end{equation}

\paragraph{Weight optimization} 
In combination, these sub-rewards reinforce factual consistency with persona facts, 
topical coherence, and language fluency.
We use their linear combination as a reward $R$ to prevent our policy from becoming overly biased towards any of the sub-rewards. For instance, while generic responses, such as \emph{``I don't know''}, have high fluency, they are discouraged by the persona-consistency sub-reward as they cannot be entailed from any persona fact. 
However, the weights must be tuned to ensure a suitable balance between the \text{sub-rewards}.
We apply grid search over the weights and choose the values that yield a policy with the best performance on a validation set (\S\ref{sec:exp3}).

\subsection{Training} 
\label{sec:actor-critic}
The goal of RL is to learn a policy, $\mathcal{P}_\theta$, for generating a response that maximizes the expected reward: 
\begin{equation}
    \mathcal{L} =  \E_{\substack{ s \in D \\ r \sim \mathcal{P}_\theta(.|s)}}{ \left[R(r,s) \right]}\,\,\text{,}
\end{equation}
where $R$ is the reward function (Equation~\ref{eq:reward}) and $s=(p,d)$ is the given persona and dialogue history that our policy has generated response $r$ for.  
Function $\mathcal{L}$ is optimized by a stochastic gradient method, where its gradient is~\cite{mnih16}:
\begin{equation}
    \frac{\partial \mathcal{L}}{\partial \theta} \!= \E_{\substack{ s \in D \\ r \sim \mathcal{P}_\theta(.|s)}}{ \!\left[ R(r,s) \frac{ \partial \log \mathcal{P}_\theta(r|s)}{\partial \theta} \right]} \!\,\,\text{.}
    \label{eq:grad}
\end{equation}
To avoid the high-variance issue, we adopt the \emph{actor-critic} method \cite{mnih16}  to \text{fine-tune} the policy function directly for our quality goals.  
This approach reduces the variance in the estimated gradient by sampling 
a single response $r \sim \mathcal{P}_\theta(.|s)$ and 
computing the difference between its reward $R(r,s)$
and the reward predicted by a critic, $\eta (t_{1..k}, s)$, for the tokens up to position $k$ in response $r$.  
The gradient in Equation~\ref{eq:grad} is then approximated as follows: 
\begin{equation}
\begin{split}
   \frac{\partial \mathcal{L}}{\partial \theta} \!\approx \! 
    \sum_k{\!(R(r,s)- \eta (t_{1..k},s)) \frac{\partial}{\partial \theta} \log \mathcal{P}_\theta(t_k|s)}\,\,\text{.}
\end{split}
\end{equation}
The critic function is $\eta = w^T h_k$, where $w$ is its trainable parameters and $h_k$ is the vector returned by the \text{TransferTransfo} model (our agent) at position $k$. 
We update the critic's parameters after each update of the policy's parameters by minimizing the squared error between its estimated rewards and the value our reward model assigns to the response: 
\begin{flalign}
    \mathcal{L}_{\eta} & = \E_{\substack{s \in D\\r \sim P_\theta(.|s)}}{\sum_k\!
    \left[ \eta (t_{1..k},s) - R(r,s)  \right]^2
    }. 
    \label{eq:loss_critic}
\end{flalign} 
\section{Experiments}
\label{sec:exp}
We measure to what extent our \text{RL-based} fine-tuning (\S\ref{sec:method}) improves the factual consistency of generated responses while retaining their semantic plausibility. 
We first introduce the corpus used in our experiments (\S\ref{subsec:dataset}). 
We then evaluate the \text{TransferTransfo-SL} and \text{TransferTransfo-RL} systems by automatic and human evaluations (\S\ref{sec:exp3}). 
We finally analyze the models we employ to estimate the factual consistency (\S\ref{sec:exp1}) and language fluency (\S\ref{sec:exp2}) sub-rewards.

\subsection{\personachat\ Corpus}
\label{subsec:dataset}
We use datasets built on the \personachat\ corpus~\cite{zhangsaizheng18},
which consists of dialogues, in English, with $6$ to $8$ turns between randomly paired human crowd-workers. 
The workers were assigned short text facts representing personas and instructed to talk to their dialogue partner \emph{naturally to discover each other's persona}. 
We chose this corpus because of its focus on promoting natural conversations while grounding conversations in the persona facts. 
Each persona consists of $4$ or $5$ facts, and on average is assigned to $8.3$ unique dialogues. 
\begin{table}[!t]
    \small
    \centering
    \begin{tabular}{lcc}
        \toprule
         & \textbf{Train} & \textbf{Validation}  \\
         \midrule 
        Num. of dialogues  & $17{,}878$ & $1{,}000$ \\
        Num. of utterances & $262{,}876$ & $15{,}602$ \\
        Num. of personas   & $955$ & $200$\\ 
        \bottomrule
    \end{tabular}
    \caption{Statistics of \personachat\ as used in the \mbox{ConvAI2} competition and our experiments.}  
    \label{tab:ps-stats}
\end{table}
We train and evaluate the aforementioned systems on the standard splits of the version of this corpus made available in ParlAI\footnote{\url{https://github.com/facebookresearch/ParlAI/tree/master/projects/personachat}} for the ConvAI2 challenge \cite{dinan19} (Table~\ref{tab:ps-stats}).  
As the test set is hidden, we evaluate the systems on the validation set. 
To create a training and evaluation sample consisting of a persona and a dialogue history (Table~\ref{tab:sample_responses}), each dialogue is split at each dialogue turn. 

\subsection{Response Generation}
\label{sec:exp3}
We study to what extent our RL approach generates a response that is factually consistent with given persona facts and semantically plausible.  
We use TranserTransfo, which performed best in automatic evaluation and second-best in human evaluation among $26$ participants in the \text{ConvAI2} competition, as a response generation model. 

\paragraph{Settings}
Following the training setup used by \newcite{wolf19}, we fine-tune 
\text{TransferTransfo}
on all training samples in \personachat\ and stop the fine-tuning after three epochs. 
We refer to this fine-tuned model as TransferTransfo-SL.  
For TransferTransfo-RL, we continue to fine-tune the TransferTransfo model with our RL approach on 90\% of the training set for one epoch, where after each policy update, the critic's parameters are updated for 5 times. 
For $R_1$, we use the BERT model trained on Dialogue NLI (\S\ref{sec:exp1}) with $\beta=2$ and for $R_3$ we use Dialogue LM (\S\ref{sec:exp2}) with $\alpha=4$. 
The maximum response length is $20$. 
The input texts are tokenized according to the GPT byte pair encoding (BPE) but the reward is computed on a completely decoded response text. 
We use the remaining $10\%$ of the training set to choose the sub-reward weights (Equation~\ref{eq:reward}) based on token-level F1-score, which indicates how well the system's responses match the 
content of human-generated responses (examined weights and their F1-scores are in Appendix~\ref{sec:weight-optimisation}), resulting in $\gamma_1=0.4$, $\gamma_2=0.16$, $\gamma_{3}=0.22$ and $\gamma_{4}=0.22$. 
The high weight of the persona consistency sub-reward ($\gamma_1$) is compatible with the goal of dialogues in \personachat , which is to reveal the persona of dialogue partners. 
The weights are also consistent with \newcite{see19}: fluency factors ($\gamma_{3}$ and $\gamma_{4}$) are more crucial than cosine-relatedness ($\gamma_{2}$) for responses in this corpus. 

\subsubsection{Automatic Evaluation}
\label{sec:auto-eval}
We evaluate these systems on the \personachat\ validation set as used in ConvAI2. 
We report PPL, F1, and BLEU to assess generated responses according to reference responses. We evaluate the factual consistency of a response and the given persona facts using our NLI model (\S\ref{sec:exp2}). It assigns inference relations between a generated response and each fact in the given persona. Given $N$ fact-response pairs in the whole evaluation set, this metric is: 
\begin{equation}
    \text{PC} = 100\frac{N_e-N_c}{N} \,\,\text{,}
\end{equation}
where $N_e$ and $N_c$ are the numbers of entailment and contradiction labels, respectively. 

\paragraph{Results}
The \text{TransferTransfo-RL} outperforms its supervised counterparts on all metrics except PPL (Table~\ref{tab:persona_consistency_result}). 
\begin{table}[!t]
    \small
    \centering
    \resizebox{0.49\textwidth}{!}{
    \begin{tabular}{@{}lcccc@{}}
        \toprule
        \textbf{Method} & \textbf{PPL} & \textbf{F1}&\textbf{BLEU}&\textbf{PC}  \\
        \midrule
         \hspace{2mm}TransferTransfo-SL & $21.31$ & $17.06$& $0.065$ & $09.32$  \\
         \hspace{2mm}TransferTransfo-RL & $22.64$ & $\mathbf{17.78}$ & $\mathbf{0.067}$ & $\mathbf{13.06}$\\
         \bottomrule
    \end{tabular}
    }
    \caption{Automatic evaluations of responses generated by TransferTransfo-SL and TransferTransfo-RL. 
    }
    \label{tab:persona_consistency_result}
\end{table}
The improvements on F1 and BLEU indicate that responses generated by \mbox{TransferTransfo-RL} are more similar to reference responses generated by humans and are not biased toward simply repeating persona facts or previous utterances. 
It also shows that responses are as informative as human-provided ones.
Our RL method decreases the average word repetition rate (Equation~\ref{eq:wr}) from $9\%$ with \mbox{TransferTransfo-SL} to $7\%$, increasing the language fluency of responses. 
So far, we observe that the RL method could retain and even improve the semantic plausibility of a response. 

\MMF[]{
Regarding the factual consistency between a response and given persona facts, \mbox{TransferTransfo-RL} scores significantly higher for the PC metric. 
This indicates that the number of  evaluation samples for which \mbox{TransferTransfo-RL} generates a response consistent with given persona facts is significantly higher than what \text{TransferTransfo-SL} does. 
}
Looking at PC in detail (Table~\ref{tab:Human_evaluation}, top), \mbox{TransferTransfo-RL} increases the frequency of cases whose generated responses are entailed from (or consistent with) persona facts by $3.41\%$ over \mbox{TransferTransfo-SL}, while reducing contradictions (or inconsistency) by $0.07\%$ and neutral by $3.61\%$; showing that fine-tuning with RL improves the policy for generating a response that is factually consistent with persona facts.  
While our combined reward function achieves good all-round performance, ablation experiments (Appendix~\ref{sec:just-ps-subreward} and \ref{sec:weight-optimisation})
show that each sub-reward is effective and necessary to capture consistency with persona facts, topical coherence, and language fluency.

\subsubsection{Human Evaluation}
\label{sec:exp4}
We also conduct a human evaluation between \mbox{TransferTransfo-RL} and \mbox{TransferTransfo-SL}.  
We randomly select \emph{100 samples}, each of which consists of a dialogue history, a  persona, and the responses generated by the examined systems. 
We ask \emph{seven human judges} (two native and five fluent English speakers)  
to assign a \emph{consistency label} from \emph{\{consistent, neutral, contradicting\}} to the response concerning the facts in the persona (instructions in Appendix~\ref{sec:instruction}). 
We also ask the human judges to rate the \emph{semantic plausibility} of
each response with an ordinal score ranging from 1 (worst) to 5 (best), 
encompassing \emph{coherence}, \emph{grammatical correctness}, and \emph{low repetitiveness}. 
\begin{table}[!t]
    \small
    \centering
    \begin{tabular}{@{}l@{}ccc@{}}
        \toprule
         & \textbf{Consistent} & \textbf{Contradicting} & \textbf{Neutral} \\
        \midrule
        \multicolumn{3}{@{}l}{\textit{Automatic Evaluation}}  \\
        TransferTransfo-SL            & $11.14$          & $01.82$           & $87.04$ \\
        TransferTransfo-RL & $\mathbf{14.81}$ & $\mathbf{01.75}$  & $\mathbf{83.43}$ \\
        $\Delta$  & $3.41 \uparrow$ & $0.07 \downarrow$&            $3.61 \downarrow$ \\
        %Human & $11.48$ & $01.54$ & $86.98$ \\
        \midrule
       \multicolumn{3}{@{}l}{\textit{Human Evaluation}} \\
        TransferTransfo-SL & $43.71$ & $17.71$& $38.58$  \\
        TransferTransfo-RL & $\mathbf{52.71}$ & $\mathbf{14.00}$ & $\mathbf{33.29}$   \\
        $\Delta$  & $9.00 \uparrow$ & $3.71 \downarrow$ & $5.29 \downarrow$ \\
         \bottomrule
    \end{tabular}
    \caption{
    Frequencies (\%) of consistency labels. Automatic evaluation uses our NLI model to assign labels for the whole evaluation set. For human evaluation, judges labeled $100$ samples. 
    $\Delta$ is the improvement of TransferTransfo-RL over TransferTransfo-SL.
    }
    \label{tab:Human_evaluation}
\end{table}
\begin{table}[!ht]
    \small
    \centering
    \begin{tabular}{lc}
        \toprule
        \textbf{Method} & \textbf{Average Semantic Plausibility} \\
        \midrule
        TransferTransfo-SL & $3.33$ \\
        TransferTransfo-RL & $\mathbf{3.50}$ \\
        \bottomrule
    \end{tabular}
    \caption{Human evaluation: semantic plausibility.} 
    \label{tab:human_fluency}
\end{table}
\paragraph{Results}
Table~\ref{tab:Human_evaluation} (bottom) shows the average percentage of consistency labels human judges assign to responses generated by \mbox{TransferTransfo-RL} and \mbox{TransferTransfo-SL}.  
The number of samples for which  \mbox{TransferTransfo-SL} generates a consistent response increases by $9\%$ using our RL fine-tuning approach while contradictions (or inconsistencies) decrease by $3.71\%$, confirming that human judges more frequently find responses generated by \mbox{TransferTransfo-RL} factually consistent with persona facts than those of \mbox{TransferTransfo-SL}. 
The number of neutral responses also decreases, suggesting fewer generic responses, as neutral responses tend to be generic \cite{welleck19}.  

Overall, Table~\ref{tab:Human_evaluation} 
shows a similar trend between the human and the automatic evaluations, 
confirming the findings of the automatic evaluation. 
Unlike the human evaluation, our automatic evaluation shows that the models generate a neutral response for most cases. 
The NLI model assesses more responses to be neutral than humans do -- humans can reason about entailment relations using their common senses, while the NLI model does not identify any relation. 
Further analysis (Appendix~\ref{sec:conf_matrix}) shows that for over half of the cases for which \mbox{TransferTransfo-SL} generates a contradicting (inconsistent) response, our \mbox{TransferTransfo-RL} generates a consistent response, indicating that the idea of using RL to fine-tune a pre-trained agent improves its capability in generating a factually consistent response with persona facts. 

In terms of semantic\MMF[ally]{} plausibility (topically coherent and linguistically fluent), Table~\ref{tab:human_fluency} shows that the human judges find responses generated by \mbox{TransferTransfo-RL} are on par with those of \mbox{TransferTransfo-SL}, showing the effectiveness of our topical coherence and fluency \text{sub-rewards}.

\subsection{Persona-Consistency Sub-reward Validation}
\label{sec:exp1}
As discussed in \S\ref{sec:method}, assessing factual consistency with persona facts can be characterized as an NLI problem. 
In this experiment, we investigate the choice of the NLI model for this sub-reward by comparing our BERT-based NLI model (\S\ref{sec:method}) with recent NLI models on the \emph{Dialogue NLI} dataset~\cite{welleck19}. 
This dataset, which is designed for evaluating factual NLI in dialogues, consists of a set of fact-utterance, fact-fact, and utterance-utterance pairs extracted from the \personachat\ corpus. 
Each pair is accompanied by a \emph{human-annotated NLI label}, i.e., entailment (or consistent), contradiction (or inconsistent), and neutral. 
Two examples of the fact-utterance pair from this dataset are:
\textit{``My dad is a priest.''} \textbf{contradicts} \textit{``Since my dad is a mechanic we had mostly car books.''};
and
\textit{``I like playing basketball''} \textbf{entails} \textit{``I prefer basketball. Team sports are fun.''}. 
This dataset contains $310{,}110$ training, $16{,}500$ validation  and $16{,}500$ test pairs. 
Besides the standard test set, which was annotated by \emph{one crowd-worker},
there is \emph{Test Gold} containing $12{,}376$ of test pairs,
which were annotated by \emph{three crowd-workers}~\cite{welleck19}. 

We compare our BERT-based NLI model with 
(1) \textbf{Majority}, which returns the majority class;   
(2) \textbf{ESIM} Enhanced Sequential Inference Model~\cite{chen-etal-2017-enhanced}, 
an LSTM-based model with inter-sentence attentions. ESIM is the state of the art on the Dialogue NLI dataset.
We use \emph{bert-base-uncased}~\cite{devlin2018bert} to encode utterances and facts. 
We fine-tune the whole model during training. 
We set the maximum input length to $128$, the learning rate to $5\times10^{-5}$, and the training- and evaluation-batch sizes to $32$ and $8$, respectively. 
We compare the NLI models using \emph{accuracy} \cite{welleck19}.  
\paragraph{Results}
\begin{table}[!t]
    \small
    \centering
    \begin{tabular}{@{}lccc@{}}
        \toprule
          \textbf{Model} & \textbf{Validation} & \textbf{Test} & \textbf{Test Gold} \\
         \midrule
         Majority & $33.33$ & $34.54$ & $34.96$ \\
         %InferSent & $85.82$ & $85.68$ & $89.96$ \\
         ESIM & $86.31$ & $88.20$ & $92.45$ \\
         \textbf{Our NLI model} & $\mathbf{86.84}$ & $\mathbf{89.50}$ & $\mathbf{93.60}$ \\
         \bottomrule
    \end{tabular}
    \caption{Accuracy of candidate NLI models for $R_1$ on the Dialogue NLI dataset.}
    \label{tab:nli-p}
\end{table}
Table~\ref{tab:nli-p} shows that the BERT-based NLI model outperforms ESIM, suggesting that our model better captures the factual relationships between an utterance and a persona fact.   
\newcite{welleck19} previously demonstrated that the performance of ESIM is sufficient to check the factual consistency between a response and persona facts; as our model outperforms ESIM, we chose our NLI model for the consistency sub-reward $R_1$. 
\MMF[]{Indeed, a more accurate NLI model reduces the noise in the reward function and consequently the errors our system makes.} 

\subsection{Response Fluency Sub-reward Validation}
\label{sec:exp2}
Sub-reward $R_{3}$ requires a language model to measure the language quality of a response. 
In this experiment, we investigate if fine-tuning a pre-trained, non-dialogue language model on dialogue utterances makes it suitable for this goal. 
To do so, we compare  
(1) \textbf{Non-Dialogue LM}, which is the GPT language model with no fine-tuning;  
and 
(2) \textbf{Dialogue LM}, which is the GPT language model fine-tuned on utterances from \personachat . 
We fine-tune the GPT language model~\cite{radford18} for three epochs on $90\%$ of utterances ($\approx\!236{,}588$) from the \personachat\ training set.
We evaluate the language model on the remaining $10\%$ ($\approx\!26{,}288$) of utterances, so the \personachat\ validation dialogues remain unseen for evaluating our dialogue systems. 
Training- and validation-batch sizes are $8$ and $16$, respectively.
Learning rate is $6.25\!\times\! 10^{-5}$, and \emph{perplexity (PPL)} is the evaluation metric.

\paragraph{Results} 
Dialogue LM substantially improves perplexity over Non-Dialogue LM (Table~\ref{tab:dlm_results}). 
This shows that the fine-tuned language model better captures the linguistic properties of dialogue utterances, yielding a more suitable language model for the fluency sub-reward $R_{3}$. 
\begin{table}[!t]
    \small
    \centering
    \begin{tabular}{ll}
        \toprule
         \textbf{Model} & \textbf{PPL} \\
         \midrule
          Non-Dialogue LM   & 108.29 \\ 
          \textbf{Dialogue LM} & \textbf{10.01} \\
         \bottomrule
    \end{tabular}
    \caption{
    The perplexity (PPL) of the language model (Dialogue LM) used for the fluency sub-reward $R_{3.1}$.  
    } 
    \label{tab:dlm_results}
\end{table}
\newcite{see19} validated the benefits of cosine similarity for estimating the coherence ($R_2$) and word repetition  ($R_{4}$) for language quality. 

\section{Discussions}
\label{sec:error-analysis}
\paragraph{Case analysis}
We presented one example of an evaluation sample in Table~\ref{tab:sample_responses},
in which  % In this example, 
the inconsistent response is generated by \text{TransferTransfo-SL} and the consistent one by \text{TransferTransfo-RL}. 
Since \mbox{TransferTransfo-SL} is fine-tuned only with reference responses and does not have any training signal for factual consistency, we speculate that variants of \emph{``I'm 50 years old''} occur in the training set leading the agent to produce a response that is inconsistent \MMF[]{with} the persona fact \emph{``I'm 40 years old''}. 
In contrast, \mbox{TransferTransfo-RL} generates a consistent response which is also topically coherent with the given question and linguistically fluent. 
\MMF[]
{
The above sample is an example of ``attribute'' consistency, where the response should express an attribute of the speaker. 
Table~\ref{tab:more_examples} shows some other evaluation samples.  
The top sample shows that TransferTransfo-RL can deal with ``have'' consistency. 
Our system correctly recognizes the number of dogs the speaker has and grounds its response on this fact.  
The evaluation sample in the middle row of Table~\ref{tab:more_examples} shows that our RL-based model can also deal with ``like-to-do'' consistency. 
}
\begin{table*}[!t]
    \small
    \centering
    \resizebox{\textwidth}{!}{
    \begin{tabular}{@{}l|l|l|l@{}}
    \toprule
\textbf{Persona} & \textbf{Dialogue History} & \textbf{TransferTransfo-SL} & \textbf{TransferTransfo-RL} \\
\midrule
\begin{tabular}{@{}lp{4cm}@{}}
    \textit{fact 1}: & i go to the gym regularly . \\
    \textit{fact 2}: & i like doing the treadmill and rowing machine . \\
    \textit{fact 3}: & i ' ve short hair . \\
    \textit{fact 4}: & two dogs live with me .  \\ 
    \textit{fact 5}: & eating is something i do when i am bored .\\ 
\end{tabular}
         & 
\begin{tabular}{@{}lp{4.1cm}}
    \textit{message}: & hello , how are you ?  \\
    \textit{response}: & hello , i am feeling fit since i visited the gym today ! how are you ?  that ' s absolutely great just fantastic after rowing and running on the treadmill , i ' m pretty hungry . do you work out ? \\
    \textit{message}: & yes after i cook all my chef food 
\end{tabular}         
         & 
\begin{tabular}{@{}p{2cm}@{}}
    i do a lot of cardio . i ' m really short , so i am always looking to
\end{tabular}                  
         & 
\begin{tabular}{@{}p{2cm}@{}}
    i love to cook , i have 2 dogs .
\end{tabular} 
\\ \midrule
\begin{tabular}{@{}lp{4cm}@{}}
    \textit{fact 1}: & my husband is adopted . \\
    \textit{fact 2}: & i like to watch nhl hockey . \\
    \textit{fact 3}: & i live is a rural farming community . \\
    \textit{fact 4}: &i ' ve a german shepherd dog .
\end{tabular}
         & 
\begin{tabular}{@{}lp{4cm}}
    \textit{message}: &the farm animals and a dog . his name is lucky . do you go to church ?  \\
    \textit{response}: & we have a german shepard named bacon . \\
    \textit{message}: & i love bacon we cook it for breakfast every sunday after church we are christain . \\
    \text{response}: & we don ' t go to church often . our religion is hockey .  \\
    \textit{message}: &  my sisters boyfriend plays hockey , she works as a pole dancer and met him there . 
\end{tabular}         
         & 
\begin{tabular}{@{}p{2cm}@{}}
    i ' m not married yet , i ' m not married yet .
\end{tabular}                  
         & 
\begin{tabular}{@{}p{2cm}@{}}
    i love to watch nhl games .
\end{tabular} 
\\\midrule
\begin{tabular}{@{}lp{4cm}@{}}
    \textit{fact 1}: & i am overweight and unhappy\\
    \textit{fact 2}: & i do not believe in god\\
    \textit{fact 3}: & i spend most of my time on facebook when i am not working\\
    \textit{fact 4}: & i work at a nursing home\\ 
    \textit{fact 5}: &my wife left me and took my children\\ 
\end{tabular}
         & 
\begin{tabular}{@{}lp{4cm}}
    \textit{message}: & hi i am a custodian named darlene , how are you ? \\
    \textit{response}: & i am good i work in a nursing home and just at home on facebook \\
    \textit{message}: & my hobbies include playing guitar and singing , do you have any?
\end{tabular}         
         & 
\begin{tabular}{@{}p{2cm}@{}}
    i do i am a drummer in a band and i work as a nurse at a nursing home
\end{tabular}                  
         & 
\begin{tabular}{@{}p{2cm}@{}}
    i like to go to church to sing with my wife 
\end{tabular} 
\\\bottomrule
    \end{tabular}
    }
    \caption{Some example personas, dialogue histories and the responses generated by our examined agents from the evaluation set. 
    }
    \label{tab:more_examples}
\end{table*}

\MMF{
Although TransferTransfo-RL outperforms TransferTransfo-SL in generating different types of consistent responses (such as `attribute'', ``have'', and ``like-to-do''),  they both struggle with generating consistent responses for evaluation samples in which understanding of persona facts and dialogue history requires common sense knowledge. 
As an example, consider the second evaluation sample shown in Table~\ref{tab:more_examples}. TransferTransfo-SL generates the response ``\emph{I'm not married yet}'' which contradicts the first fact of the given persona ``\emph{My husband is adopted.}''; it seems the model does not have enough knowledge to capture the semantic relationship between ``\emph{my husband}'' and ``\emph{marriage}''.
The bottom evaluation sample in Table~\ref{tab:more_examples} demonstrates the lack of common sense knowledge for TransferTransfo-RL as well. 
The response ``\emph{I like to go to church to sing with wife}'' contradicts the fact ``\emph{My wife left me and took my children}'' in the given dialogue history. 
}

\paragraph{Limitations}
One limitation of our work is to narrow a speaker's persona to a set of facts expressed as short sentences.  
Persona has other aspects, such as speaking styles, which need a separate study.  
Nevertheless, the research question and experiments presented in this work demonstrate the benefits of RL methods for fine-tuning transformer-based models, which are already pre-trained, to obtain a policy more aligned with target quality factors. 
Other aspects of the persona can also be involved in the reward function\MMF[]{,} given that our method potentially reduces the need for the high-quality demonstration responses generated by humans for supervised fine-tuning.  

\paragraph{Future directions}
In this paper, we  demonstrate the \MMF[advantages]{effectiveness} of RL over SL for fine-tuning pre-trained neural models (like GPT) for generating responses \MMF[which]{that} fulfill  quality goals such as factual consistency with given persona facts and semantic plausibility in a single round of dialogue.  
Therefore, the next step might be adopting our reward function to generate factually-consistent responses while retaining the diversity of responses through  multiple rounds of dialogue.

\section{Related Work}
\label{sec:rel}
There are two types of approach to persona consistency. 
The first category includes systems that learn speaker-level embeddings from responses  produced by a particular speaker \cite{Jiweili16persona,madotto19}. 
These systems depend on the availability of suitable responses performed 
by the speaker whose persona we wish to imitate. 
If those \MMF[dialogues]{responses} do not reveal the persona information, dialogue systems cannot learn the persona. 
Moreover, these systems cannot be adapted to new personas at deployment time since the persona embeddings must be learned from training data. 
So our approach is complementary to them and not directly comparable. 

The second category includes systems that rely on a set of facts about a persona.
For example, \newcite{zhangsaizheng18} propose a key-value memory neural model for this task. 
This model is outperformed by TransferTransfo~\cite{wolf19}, which is used in our experiments. 
\newcite{welleck19} rank a given set of utterances using an NLI model to select a persona-consistent response.
In contrast, we use NLI to train a generative model. 
\newcite{songhaoyu20} propose an NLI-based reward for persona consistency that calculates a score using only the persona facts with the highest entailment and contradiction probabilities, 
rather than the whole persona.  
Their approach does not reward topical coherence,
which we found crucial for relieving effects of the persona-consistency sub-reward on the quality of response. 

Persona consistency was also a quality target in the \text{ConvAI2} dialogue generation competition~\cite{dinan19}. 
The winner of the human evaluation part of ConvAI2
is the ``Lost in Conversation'' system \cite{dinan19}, which is also a transformer-based model
trained by SL on two extra datasets besides PersonaChat. 
In our paper, we used TransferTransfo trained only on PersonaChat. 
Our experiments showed that our idea of using RL for fine-tuning neural agents improves factual consistency between a response and persona facts by accounting for it in its reward function. 

RL has been extensively used for training task-oriented dialogue systems (e.g., \newcite{nogueira17,liubing18}). 
Unlike task-oriented scenarios, where a reward can measure if a task is fulfilled or not, incorporating persona\MM[s]{} facts lacks a straight-forward measurable outcome.  
\newcite{lijiwei16} use RL for generating open-domain dialogue using REINFORCE (instead of Actor-Critic) and an RNN-based model. 
This agent has no notion of factual consistency with facts about a persona, 
so is not comparable with our system. 

\section{Conclusions}
We proposed to fine-tune response generation models by RL to improve on the quality goals that matter, e.g., factual consistency between a response and persona facts while retaining semantic plausibility.
We adopted the actor-critic method for fine-tuning a pre-trained transformer-based model by defining an efficient and effective reward function measuring persona consistency, topical coherence, and language fluency.  
Automatic and human evaluations on \personachat\ demonstrate that 
compared to just using supervised learning,
further fine-tuning with RL yields %
responses that are more frequently factually consistent with persona facts while still semantically plausible. %

\section*{Acknowledgments}
\MMF[]{
This work was supported by the German Research Foundation through the German-Israeli Project Cooperation (DIP, grant DA 1600/1-1 and grant GU 798/17-1). 
We thank Kevin Stowe and Leonardo Filipe Rodrigues Ribeiro for valuable feedback on earlier drafts of this paper. 
We also thank anonymous reviewers for their constructive suggestions. 
}

% \section*{Acknowledgments}
% This work was supported by the German Research Foundation through the German-Israeli Project Cooperation (DIP, grant DA 1600/1-1 and grant GU 798/17-1). 

% We thank Ji-Ung and  Leonardo Filipe Rodrigues Ribeiro, and Johannes Dexenberg for their valuable feedback on earlier drafts of this paper. 

% We also thank anonymous reviewers for their useful suggestions for improving the quality of the paper. 

\bibliography{lit,lit_paper}
\bibliographystyle{acl_natbib}

\clearpage
\appendix

\section{Using Only the Persona-Consistency Sub-reward $R_1$}
\label{sec:just-ps-subreward}
By using the persona-consistency sub-reward as the only reward function, the policy learns only to generate responses that state information about the persona facts and ignores the topical coherence with its former utterance. 
\begin{table}[!ht]
    \small
    \centering
   
    \begin{tabular}{@{}l@{}}
         \toprule
         \textbf{Persona}  \\
         \midrule 
         \begin{tabular}{@{}p{7.5cm}@{}}
            \textit{fact 1}: i prefer vinyl records to any other music  format. \\
            \textit{fact 2}: i fix airplanes for a living. \\
            \textit{fact 3}: i drive junk cars that no one else wants. \\ 
            \textit{fact 4}: i think if i work hard enough i can fix the world. 
        \end{tabular} 
        \\
        \midrule 
        \textbf{Dialogue}  \\
        %\midrule
        \begin{tabular}{@{}p{7.5cm}@{}}
            \textit{message}: hi how are you doing?\\
            \textit{response}: i am fine. i am a mechanic, so i fix airplanes. i fix them.  \\
            \textit{message}: do you have any hobbies? \\
            \textit{response}: i enjoy listening to vinyl records. i also fix planes. i also fix them. i also fix \\
            \textit{message}: sounds cool. \\ 
            \textit{response}:  i also fix airplanes. i also fix airplanes. i also fix them. i also fix them.
        \end{tabular}
\\
\bottomrule
\end{tabular}
\caption{Response generated by the TransferTransfo model where it is fine-tuned by only the persona-consistency sub-reward ($R=R_1$).}
\label{tab:just_ps_subreward}
\end{table}

Table~\ref{tab:just_ps_subreward} illustrates an example dialogue that is conducted with an agent trained with only the persona consistency sub-reward ($R=R_1$). 
The agent always repeats, \emph{``i fix airplanes. i fix them.''}, no matter what the input message is about. 
This problem not only produces topically irrelevant responses but also makes the agent look nagging and \text{self-centered} in a conversation. 
\begin{table}[!ht]
    \small
    \centering
    \begin{tabular}{@{}l@{}}
         \toprule
         \textbf{Persona}  \\
         \midrule 
         \begin{tabular}{@{}p{7.5cm}@{}}
            \textit{fact 1}: i like hunting. \\
            ...
        \end{tabular} 
        \\
        \midrule 
        \textbf{Dialogue}  \\
        \begin{tabular}{@{}p{7.5cm}@{}}
            \textit{message}: hi how are you doing?\\
            \textit{response}: hunting hunting hunting hunting hunting hunting \\
            \textit{message}: do you have any hobbies? \\
            \textit{response}: hunting hunting hunting hunting hunting hunting \\
        \end{tabular}
\\\bottomrule
\end{tabular}
\caption{Another example dialogue with an agent that is trained  by only the persona-consistency sub-reward ($R=R_1$).}
\label{tab:just_ps_subreward_hunting}
\end{table}

Table~\ref{tab:just_ps_subreward_hunting} illustrates another example dialogue with the agent where it is trained only by persona-consistency sub-reward. 
The agent keeps repeating \emph{``hunting''} from the persona to maximize its reward. 
The NLI model used for $R_1$ evaluates the inference relation between a response and a persona and does not capture the topical coherence of the response with its former utterance and language fluency of the response.  
It is therefore necessary to use $R_1$ in combination with topical coherence ($R_2$) and language fluency sub-rewards ($R_3$ and $R_4$), as we propose in our reward function.   
\section{Weight Optimization and Reward Ablation}  
\label{sec:weight-optimisation}
We examine various weight sets $(\gamma_1, \gamma_2, \gamma_3, \gamma_4)$ to balance the contribution of sub-rewards in the complete reward function on the held out set (10\% of the \personachat\ training set). 
Table~\ref{tab:examined_weights} shows those weights. 
\begin{table}[!ht]
    \small
    \centering
    \begin{tabular}{ccccc}
        \toprule
        $\mathbf{\gamma_1}$ & $\mathbf{\gamma_2}$ & $\mathbf{\gamma_3}$ & $\mathbf{\gamma_4}$ & $\mathbf{F_1}$\\ \midrule
0.00 & 0.00 & 0.00 & 0.00 &  02.04 \\ 
1.00 & 0.00 & 0.00 & 0.00 &  16.34 \\ 
0.00 & 1.00 & 0.00 & 0.00 &  16.12 \\ 
0.00 & 0.00 & 1.00 & 0.00 &  12.77 \\ 
0.00 & 0.00 & 0.00 & 1.00 &  17.91 \\ \midrule
0.70 & 0.00 & 0.30 & 0.00 &  16.37 \\ 
0.65 & 0.00 & 0.35 & 0.00 &  19.90 \\ 
0.60 & 0.00 & 0.40 & 0.00 &  16.98 \\ 
0.50 & 0.00 & 0.50 & 0.00 &  16.07 \\ 
0.25 & 0.25 & 0.25 & 0.25 &  18.27 \\ 
0.60 & 0.20 & 0.00 & 0.20 &  15.54 \\ 
0.40 & 0.20 & 0.20 & 0.20 &  20.57 \\ 
\textbf{0.40} & \textbf{0.16} & \textbf{0.22} & \textbf{0.22} &  \textbf{20.75} \\ 
0.45 & 0.13 & 0.17 & 0.20 &  19.98 \\ 
0.47 & 0.12 & 0.17 & 0.20 &  19.95 \\ 
0.47 & 0.10 & 0.17 & 0.21 &  20.33 \\ 
0.47 & 0.10 & 0.19 & 0.19 &  19.73 \\ 
0.50 & 0.10 & 0.16 & 0.20 &  19.10 \\ 
%0.40 & 0.16 & 0.22 & 0.22 &  19.56 \\ 
0.40 & 0.20 & 0.15 & 0.20 &  19.34 \\ 
0.43 & 0.20 & 0.12 & 0.20 &  20.22 \\ 
0.45 & 0.20 & 0.12 & 0.20 &  20.13 \\ 
0.45 & 0.25 & 0.00 & 0.25 &  17.40 \\ 
0.40 & 0.10 & 0.25 & 0.25 &  18.93 \\ 
0.40 & 0.15 & 0.20 & 0.20 &  19.80 \\ 
0.40 & 0.20 & 0.20 & 0.15 &  20.44 \\ 
0.45 & 0.17 & 0.21 & 0.17 &  18.85 \\ 
0.50 & 0.15 & 0.15 & 0.15 &  20.25 \\ 
0.47 & 0.13 & 0.20 & 0.15 &  19.97 \\ 
0.50 & 0.15 & 0.20 & 0.15 &  17.64 \\ 
0.55 & 0.15 & 0.15 & 0.10 &  19.15 \\ 
\bottomrule
    \end{tabular}
    \caption{The examined sub-reward weights and their corresponding F1 on our validation set, i.e., 10\% of the \personachat\ training set.}
    \label{tab:examined_weights}
\end{table}
The balanced weights give the highest F1 score, suggesting that a combination 
of sub-rewards leads to responses that are more similar to the human responses. 

We also evaluate the use of each sub-reward in isolation, and show the results in Table \ref{tab:ablation}, in comparison with 
our chosen balanced weights in the bottom line. 
\begin{table*}[!ht]
    \small
    \centering
    \resizebox{\textwidth}{!}{
    \begin{tabular}{cccc|cccccccc}
        \toprule
        $\mathbf{\gamma_1}$ & $\mathbf{\gamma_2}$ & $\mathbf{\gamma_3}$ & $\mathbf{\gamma_4}$ &  $\mathbf{F_1}$& $\mathbf{PPL}$ & \textbf{Repetition(\%)} & \textbf{Consistent (\%)} & \textbf{Neutral (\%)} & \textbf{Contradiction (\%)} & \textbf{PC} \\ \midrule
        1.00 & 0.00 & 0.00 & 0.00 &  16.34 & 25.35 & 36.20 & \textbf{34.02} & 64.62 &  01.36 & \textbf{66.33}\\ 
        0.00 & 1.00 & 0.00 & 0.00 &  16.12 & 21.44 & 13.36 & 12.24 & 86.34 &  \textbf{01.42} & 55.50\\ 
        0.00 & 0.00 & 1.00 & 0.00 &  12.77 & \textbf{04.76} & 12.21 & 00.46  &  \textbf{99.35} &  00.20 & 51.49\\ 
        0.00 & 0.00 & 0.00 & 1.00 &  17.91 & 21.35 & \textbf{01.62} & 03.53 &  95.90 &  00.57 &  50.13\\ 
        \midrule
        0.40 & 0.16 & 0.22 & 0.22 &  \textbf{20.75} & 12.36 & 09.61 & 14.14 & 84.54   & 01.32  & 56.50\\ 
\bottomrule
    \end{tabular}
    }
    \caption{Performance metrics on our validation set (10\% of the \personachat\ training set) when training is performed with each individual sub-reward and our chosen weighted sum of sub-rewards. }
    \label{tab:ablation}
\end{table*}
For the other metrics, we can see that $\gamma_1=1$ maximizes the number of 
entailments from the persona facts, $\gamma_3=1$ minimizes perplexity, and
$\gamma_4=1$ gives lowest repetition. 
Besides F1 score, the balanced weights give good performance across perplexity, repetition, and persona consistency. 
The setups with fewer neutral responses also tend to have more responses that contradict the persona facts, e.g., for $\gamma_1=1$. 
Neutral responses are a trivial way to avoid contradictory responses and
the setup with the least contradictions, $\gamma_3=1$, 
has almost no responses that are consistent with the persona facts. 
The better overall persona consistency is reflected in the highest PC score for $\gamma_1=1$ and next highest for the balanced weights, which trades of PC for less repetition, lower perplexity and a higher F1 score.

\section{REINFORCE vs Actor-Critic}
\label{sec:reinf_vs_ac}
\begin{figure}[!ht]
    \centering
    \begin{tabular}{c}
        \includegraphics[scale=0.14]{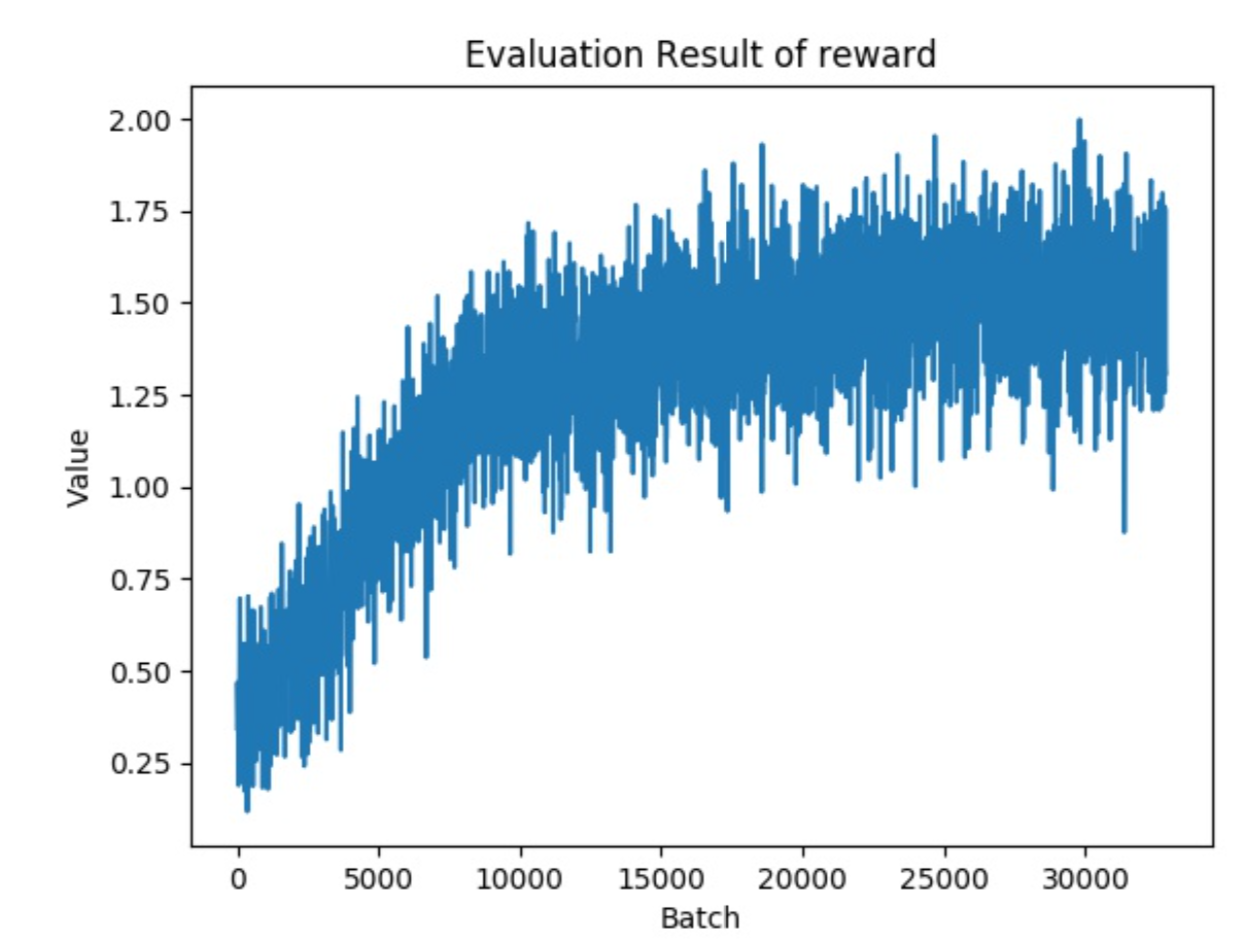}  
    \end{tabular}
    \caption{The reward curve during training by REINFORCE.}
    \label{fig:reinforce}
\end{figure}

\begin{figure}[!ht]
    \centering
    \begin{tabular}{l}
     \includegraphics[scale=0.14]{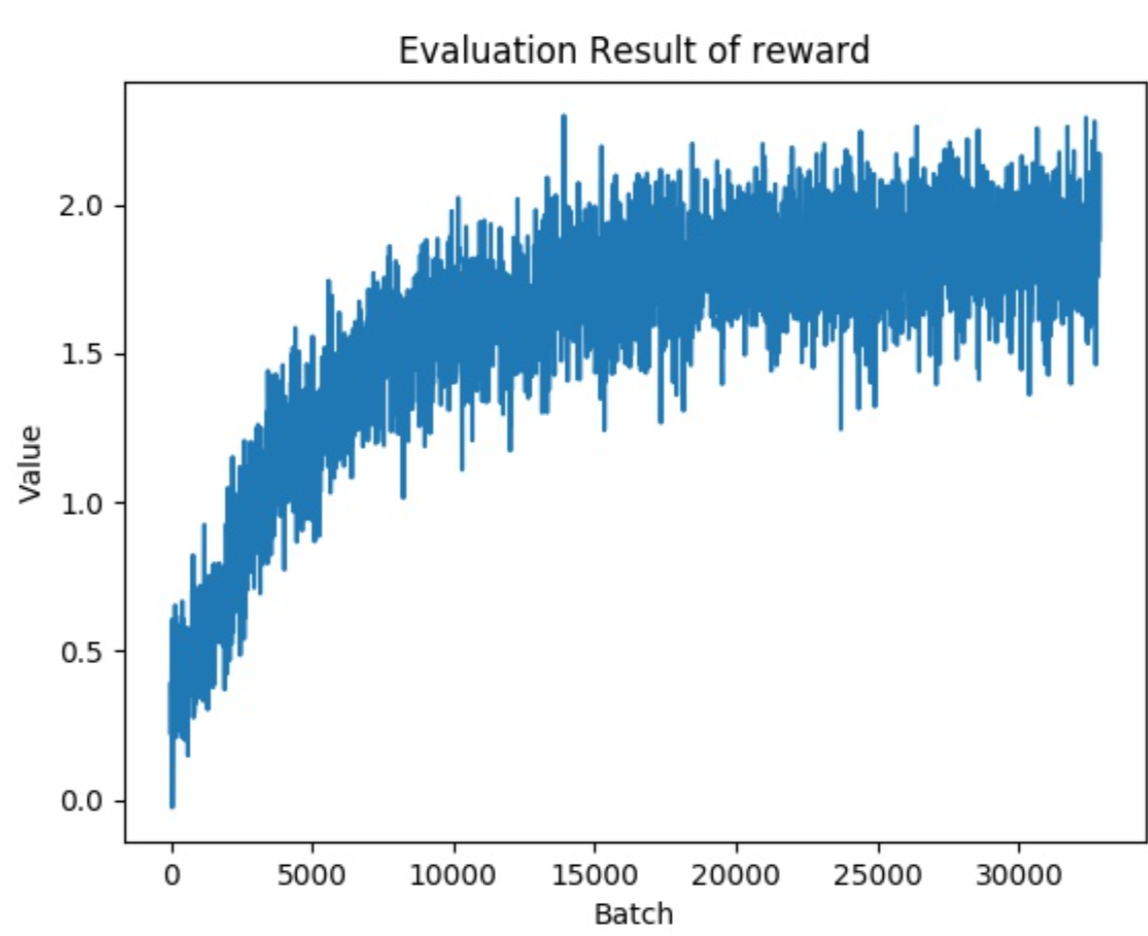} 
    \end{tabular}
    \caption{The reward curve during training by \mbox{Actor-Critic}.}
    \label{fig:ac}
\end{figure}

Figures~\ref{fig:reinforce} and ~\ref{fig:ac} show the trend of changes in our reward function during training by REINFORCE and Actor-Critic, respectively. %
All parameters are the same for the two experiments. 
%
%In the actor-critic model, critic training is performed for five epochs.
%
We observe that the actor-critic approach converges faster \MMF[as well as]{and} also is less noisy (has a lower variance) than REINFORCE. 

\section{Human Evaluation}
\label{sec:instruction}
For each sample, we show to each participant a set of persona facts, a dialogue history, and the response generated by one of \text{TransferTransfo-SL} and \text{TransferTransfo-RL}.   
We instruct our participants to assess semantic plausibility according to the following objective definition: \emph{``grammatical correctness, lowest repetitiveness, and coherence''}. 
The plausibility rates are integer values between 1 and 5, where 5 is most plausible. 

To measure persona consistency, we instruct participants as follows: 

An answer is considered consistent if:
    \begin{itemize}
        \item It contradicts with neither the dialogue history nor the persona facts;
        \item It is relevant to any of the given persona facts. 
    \end{itemize}
An answer is considered neutral if:
    \begin{itemize}
        \item It contradicts with neither the dialogue history nor the persona facts;
        \item It is \textbf{not} relevant to any of the given persona facts. 
    \end{itemize}

\section{Human Evaluation: Confusion Matrix}
\label{sec:conf_matrix}
Table~\ref{tab:human_evaluation_confusion} 
presents the distributions of consistency labels for \text{TransferTransfo-RL}'s responses 
given the consistency labels for \text{TransferTransfo-SL}'s responses.
For the majority of cases whose \text{TransferTransfo-SL}'s responses are contradictory or neutral, 
\text{TransferTransfo-RL} generates consistent responses, showing improved factual consistency with persona facts. 
However,  \text{TransferTransfo-RL} generates contradictory responses for some
cases whose \text{TransferTransfo-SL} responses are consistent with their personas. 
This may be due to errors in the NLI model's predictions of entailment, 
hence a more accurate NLI model may improve the quality of the reward function and consequently the consistency of responses. Alternatively, these contradictory responses may receive
high rewards from the topic consistency and fluency sub-rewards, which
could override $R_1$.

\begin{table}[!t]
    \small
    \centering
    \setlength{\tabcolsep}{3pt}
    \begin{tabular}{llccc}
        \toprule
                         & & \multicolumn{3}{c}{\textit{TransferTransfo-RL label}} \\
         & & Consistent & Neutral & Contradicting \\
        \midrule
        \textit{Transfer} & Consistent   &  $57.46$  & $27.73$  & $14.82$\\
        \textit{Transfo} & Neutral      &  $46.87$ &  $44.99$ & $08.14$\\
        \textit{-SL} & Contradicting&  $52.48$ & $22.05$  & $25.47$\\
         \bottomrule
    \end{tabular}
    \caption{
    Each row corresponds to the cases in the human evaluation for which \text{TransferTransfo-SL} received a particular consistency label. The values in each row show the percentages of
    consistency labels for \text{TransferTransfo-RL} for the same data points.}
    \label{tab:human_evaluation_confusion}
\end{table}

\end{document}